\documentclass{article}
\usepackage[utf8]{inputenc}
\usepackage[english]{babel}
\usepackage{caption}
\usepackage{float}
\usepackage{subcaption}

\usepackage[lmargin = {2.7cm},
rmargin = {2.7cm},
tmargin = {2.2cm},
bmargin = {2.2cm}
]{geometry}

\usepackage{comment}
\usepackage[hidelinks]{hyperref}
\usepackage[capitalise]{cleveref}

\usepackage[
backend=biber,
style=numeric,
sorting=nty
]{biblatex}
\usepackage{graphicx}
\graphicspath{ {./} }

\title{OWL Reasoners still useable in 2023}
\author{\textbf{Konrad Abicht} \\ k.abicht@gmail.com}
\date{13.09.2023}

\begin{document}

\maketitle

\begin{abstract}
    In a systematic literature and software review over 100 OWL reasoners/systems were analyzed to see if they would still be usable in 2023. This has never been done in this capacity. OWL reasoners still play an important role in knowledge organisation and management, but the last comprehensive surveys/studies are more than 8 years old. The result of this work is a comprehensive list of 95 standalone OWL reasoners and systems using an OWL reasoner. For each item, information on project pages, source code repositories and related documentation was gathered. The raw research data is provided in a Github repository for anyone to use.
\end{abstract}
\section{Introduction}

There are many surveys and studies concerning OWL reasoners. Some examine the underlying methods and functionality, others compare performance metrics. One might think that the field of OWL reasoners is well established and that there is software for each relevant application. But this is not the case. Instead I have noticed that well known reasoners have hardly been updated in the last 10 years (e.g. HermiT). Some are still usable, mostly as Protégé plugins, but it raises the question whether new (research or commercial) projects should rely on them. How are they maintained? Are bugs detected and dealt with? Do projects maintain their software dependencies? People interested in OWL reasoners today face many obstacles. To get a neutral view on the software landscape, I conducted a survey between May and July 2023. You hold the results of this work in your hands.

This paper is structured as follows: Section 2 contains short summary of required background knowledge. Section 3 then summarises related work. Section 4 describes my methodology and the section 5 presents results of my research. Finally, in section 6, I draw my conclusions and in section 7, I provide further starting points for future work.

\subsection{Publicly available research data}

All research data is publicly available via a Github repository. It contains a CSV file with a list of analyzed OWL reasoners as well a CSV file with systems using a foreign OWL reasoner. For each entry there is metadata about installation, usability and references such as source code repository. All this data is available at the following URL:

\begin{center}
    https://github.com/k00ni/owl-reasoner-list
\end{center}

I invite everyone to contribute. The repository is designed in a way to support further research and additions, so that others can continue the work in the years to come without having to start from scratch each time.

\begin{figure}[H]
    \centering
    \begin{minipage}{.4\textwidth}
        \centering
        \includegraphics[scale=0.2]{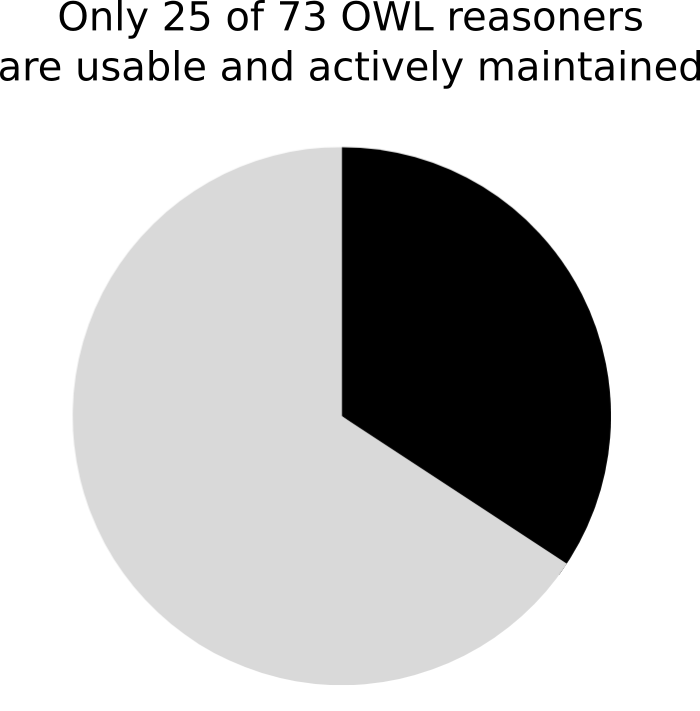}
        \captionof{figure}{}
        \label{fig:test1}
    \end{minipage}
    \begin{minipage}{.4\textwidth}
        \centering
        \includegraphics[scale=0.2]{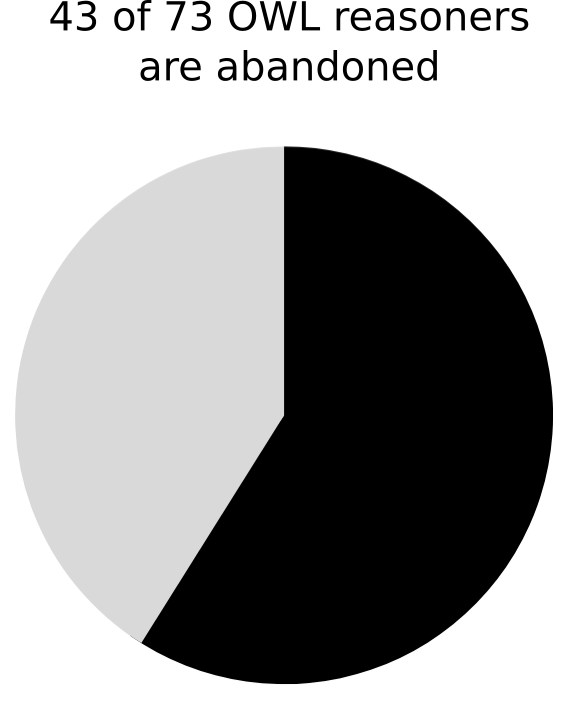}
        \captionof{figure}{}
        \label{fig:test2}
    \end{minipage}
\end{figure}

\section{Reader background}

You should have an extended knowledge of Semantic Web technologies and concepts such as RDF, RDFS, OWL 1/2 and OWL reasoning. There are many programming/software environments used to develop OWL reasoners, so basic knowledge in compiling and executing programs is recommended. Basic knowledge of software development using distributed version control systems, such as Git, is helpful. Below is a brief summary of the most widely used systems.

\subsection{Protégé}

Protégé\cite{sivakumar2011ontology} is an ontology editor well known to ontologists and Semantic Web developers. It has been developed by Stanford University\footnote{https://protege.stanford.edu/}. It provides tools for developing and maintaining OWL ontologies. There are many plugins available, for instance to use an OWL reasoner. Protégé is written in Java and runs on Windows 10/11 as well as Ubuntu Linux.

\subsection{OWL API}

OWL-API \cite{horridge2011owl} is written in Java and provides an Application Programming Interface for managing OWL ontologies. In addition to parsing and manipulating OWL ontologies, it also allows the use of reasoners. It also includes validators for different OWL profiles, for instance OWL 2 QL\footnote{\url{https://www.w3.org/TR/owl2-profiles/\#OWL\_2\_QL}}, OWL 2 EL\footnote{\url{https://www.w3.org/TR/owl2-profiles/\#OWL\_2\_EL}} or OWL 2 RL\footnote{\url{https://www.w3.org/TR/owl2-profiles/\#OWL\_2\_RL}}. Further information and source code can be found on the project page\footnote{https://owlcs.github.io/owlapi/}.

\section{Related work}

Since the publication of OWL in 2001, there have been many benchmarks and surveys comparing and evaluating OWL reasoners. In the following only the most recent and relevant ones are presented.

The most recent and relevant publication \cite{lam2023performance} is from 2023. The authors evaluated the performance of six prominent OWL 2 DL compliant reasoners (such as Pellet, FaCT++ and Hermit) on various reasoning tasks. One of their findings was that many projects are no longer actively maintained. This supports my results and observations, even though their metrics differ from the ones used in this paper (they used a wider range for activity: last 10 years).

The website http://owl.cs.manchester.ac.uk/tools/list-of-reasoners/ is often cited in publications. It contains a list of 39 reasoners, each entry having a summary and some metadata such as supported interfaces. Last update was in June 19, 2018. Authors behind it are Uli Sattler and Nico Matentzoglu. They link to https://www.w3.org/2001/sw/wiki/OWL/Implementations, an overview of OWL implementations such as reasoners, editors and APIs. It contains fewer entries, but still has some interesting details per entry. It was last updated on June 9, 2020.

In \cite{khamparia2017comprehensive} (2017) the authors report on a survey of many Semantic Web tools and services, including a comparison of 23 OWL reasoners in terms of their features. All reasoners were reported as usable, but unfortunately there was no information about maintenance status of OWL reasoners.

The authors of \cite{parsia2017owl} performed a reasoner evaluation in 2015 using 13 reasoners (such as FaCT++, HermiT and jcel). They reported that all the reasoners were usable, although they performed very differently in the competition (e.g. performing DL consistency checking or DL classification). Also in 2015 a survey was conducted using 35 OWL reasoners \cite{matentzoglu2015survey} by the same authors who also created http://owl.cs.manchester.ac.uk/tools/list-of-reasoners/. In the survey they asked developers about functionality, language used and for feedback for recommended usage. They report that eight reasoners are barely maintained and CEL and HermiT are not maintained at all.

It is important to note that there no survey or benchmark has been found that compares more than 35 reasoners. They either focus on the most prominent reasoners or only use reasoners for a certain OWL profile. Some surveys also include other systems, such as editors or IDEs, which are out of the scope of this work.

\section{Methodology}

The motivation behind this survey was to get an objective view of available OWL reasoners. For a comprehensive overview information such as project status and usability is important. This information needs to be researched, ordered and assessed for each OWL reasoner.

The following list contains major research questions:

\begin{enumerate}
    \item Which OWL reasoners are available?
    \item Which are still usable?
    \item Which are still actively maintained?
\end{enumerate}

I have chosen these research questions in order to get as neutral an overview as possible, they provide a basis for further investigation. The second question refers to usable tools, which is crucial because if a software can no longer be started, it is useless. In the following I specify the terms OWL reasoner, usable and actively maintained.

\subsection{Important terms}

\subsubsection{OWL Reasoner}

An OWL reasoner (or semantic reasoner) is basically an inference machine that, infers logical consequences from a given set of axioms (RDFS/OWL data). However, my analysis has shown, that OWL reasoners can have a wide range of features. For example, checking the consistency of an ontology, or checking whether a given set of rules applies. In this survey, the range of features scope doesn't matter as long as it provides OWL reasoning in some way.

An OWL reasoner was included in the analysis if it is an open source project or is provided free of charge. Furthermore,  only software released after the public announcement of OWL 1 in 2004\footnote{According to https://www.w3.org/TR/owl-features/} was considered.

\subsubsection{Usable}

An OWL reasoner (or system using a foreign OWL reasoner) is considered usable, if it meets the following criteria:

\begin{enumerate}
    \item The OWL reasoner can be started successfully on Ubuntu 20.04 (64-bit, X11). Or using PlayOnLinux\footnote{https://www.playonlinux.com/en/} + Wine\footnote{https://www.winehq.org/}, if software requires a Windows environment. My local machine hardware details are: IBM ThinkCentre with 8 GB RAM, Intel Core i5-7400T and solid state disk.
    \item I did not test on my local machine if I did not know the development environment of the software. Systems such as Haskell or Prolog require a certain software stack, that can only be set up properly with prior knowledge. This is also the case, if a project was only available as a Java API or Protégé plugin. No custom code was written and executed to test the reasoner. In all these cases I have instead examined project documentation and projects/publications, that use a particular reasoner to determine, whether it is usable or not.
\end{enumerate}

In some cases it was not possible to clarify whether an OWL reasoner was usable without any doubt. Relevant information has been added and its up to the reader to conduct further analysis. This approach was taken to avoid misunderstandings and false claims.

\subsubsection{Actively maintained}

Active maintenance is an important criteria for people looking for software for a specific use case. However, it is difficult to objectively assess whether a software is still being actively maintained or not. The following aspects have been chosen because they are objective and rooted in software engineering.

\begin{enumerate}
    \item An open source OWL reasoner is considered to be actively maintained, if the last release/code commit is not older than 3 years or there are other activities by the developers/maintainers, such as responses to bug reports in the same time period.
    \item A closed source OWL reasoner is considered to be actively maintained, if the developer has provided software at the time of the survey.
\end{enumerate}

Why 3 years? In my experience open source projects can be inactive for 2-3 years, but still receive important updates from time to time. 3 years seemed to be reasonable enough for this type of survey.

\subsection{Literature- and Internet research}

The OWL reasoner has its roots in the scientific community. Therefore literature research was the starting point. I used Google Schoolar\footnote{https://scholar.google.com/} to find relevant publications. Initially, I searched for "OWL Reasoner", but refined my search by using specific project names with the term reasoner to avoid ambiguous results. Only publicly available publications were considered, which means that if a publication was behind a paywall, it was ignored. A paywalled publication may contain valuable information, but I found it unreasonable to pay 30 EUR or more just to gain access.

An OWL reasoner was selected for further research, if there was a scientific publication or technical report about it, or it had a dedicated project repository. Small demos were ignored, because their usability in the long term was questionable. Systems that provide reasoning services using a third-party OWL reasoner were managed separately. They were included because they provide a benefit to users for certain applications, such as selecting an appropriate reasoner for a given ontology (meta reasoner).

All OWL reasoners were collected in a CSV file, supplemented by meta data such as usability, maintenance status or project website. This list provided a stable overview of available OWL reasoners and also helped to find relations between some projects. At https://github.com/k00ni/owl-reasoner-list you can get all raw data, such as the mentioned CSV file, which was created during this analysis.

In addition to the literature search, a separate internet search was carried out for each OWL reasoner using Google Search. Project websites, source code repositories and other relevant information were of interest. One note: Some projects were mentioned in an article and only have a project page, but there was no dedicated article about them.

Github\footnote{https://www.github.com} is one of the main platforms for finding open source projects\footnote{https://www.linuxfoundation.org/blog/hosting-open-source-projects-on-github-nine-things-you-need-to-know}. Therefore a search using reasoner name (and sometimes the term "reasoner" too) was conducted. The same applies for projects on other platforms such as Bitbucket or Sourceforge.

\subsection{Software review and analysis}

Each OWL reasoner was examined. First, executable binaries, if available, were downloaded and tested on my local machine running Ubuntu 20.04. After downloading a binary, I tried to find information about installation/setup. Based on the available information and my personal knowledge the binary was installed/setup and executed.

If the software started and showed a CLI or user interface the test was complete. If a library (e.g. as Jar-file\footnote{\url{https://en.wikipedia.org/wiki/Jar\_(file\_format)}}) was provided by the project, I tried to determine usability by searching the project files and the web. For example, if a library was provided and the project used a continuous integration pipeline that did not report any errors during the last run, it can be assumed that the library still works.

If no usable files could be found, no further investigation was conducted. No custom code was developed to test a reasoner.

\section{Results}

The structure of this section is as follows. First, a short overview about the most important statistical figures related to OWL reasoners (Table \ref{table-stat-figures-reasoners}) and systems using a foreign OWL reasoner (Table \ref{table-stat-figures-systems}) is given.

This is followed by a list of available OWL reasoners. Each block consists of a summary, download and repository part as well as an installation and usage part. This is followed by a list of systems which use a foreign OWL reasoner. Finally, a list of (probably) unusable OWL reasoners is given. Table \ref{table-usable-maintained-owl-reasoners} in the appendix shows a list of all usable and maintained reasoners. Table \ref{table-all-owl-reasoners} contains a list of all analyzed OWL reasoners. A more detailed overview can be found in the mentioned Github repository (look for the files: "reasoner.csv" and "system-using-foreign-reasoner.csv").

\begin{table}[H]
    \centering
    \begin{tabular}{|l|l|}
        \hline
        \textbf{The amount of ...}                          &                \\ \hline
        ... usable and maintained OWL reasoners is          & 25             \\ \hline
        ... usable OWL reasoners (maintenance ignored) is   & 32             \\ \hline
        ... maintained OWL reasoners (usability ignored) is & 28             \\ \hline
        ... OWL reasoners that weren't tried is             & 9              \\ \hline
        ... OWL reasoners with no available files is        & 20             \\ \hline
    \end{tabular}
    \caption{Overview of figures for a total of \textbf{73} analyzed OWL reasoners \label{table-stat-figures-reasoners}}
\end{table}

\begin{table}[H]
    \centering
    \begin{tabular}{|l|l|}
        \hline
        \textbf{The amount of systems using a foreign OWL reasoner that ...} &   \\ \hline
        ... are usable and maintained is                                     & 4 \\ \hline
        ... are usable (maintenance ignored) is                              & 8 \\ \hline
        ... are maintained (usability ignored) is                            & 5 \\ \hline
        ... weren't tried is                                                 & 4 \\ \hline
        ... are have no available files is                                   & 9 \\ \hline
    \end{tabular}
    \caption{Overview of figures for a total of \textbf{22} analyzed systems using a foreign OWL reasoner \label{table-stat-figures-systems}}
\end{table}

\break

\subsection{Usable OWL Reasoners}

Below is a list of all usable OWL reasoners.

\subsubsection{AllegroGraph}

\textbf{Summary:} AllegroGraph\footnote{https://allegrograph.com/products/allegrograph/} is an RDF graph database which provides data storage and other services. One of these is an RDFS++ reasoner, which supports all RDFS predicates and also some OWL predicates.\footnote{https://franz.com/agraph/support/documentation/current/agraph-introduction.html\#reasoning-intro}.
Source code is not available. The most relevant publication is from Diogo Fernandes and Jorge Bernardino \cite{fernandes2018graph}. The company behind AllegroGraph has published many white papers about this reasoner, but no information was found about their RDFS++ Reasoner\footnote{https://allegrograph.com/white-papers/}.

\textbf{Download and Repository:} AllegroGraph can be in version 7.3.1 downloaded for different Linux systems (Ubuntu, RHEL,...)\footnote{https://franz.com/agraph/downloads/}. You can also use a Docker container to run it. There was no binary available for Windows. The AllegroGraph client source code is available on Github in several languages\footnote{https://github.com/franzinc?tab=repositories}.

\textbf{Installation and Usage:} Installation introductions are available on the download page. Usage information can be found at the following link\footnote{https://franz.com/agraph/support/documentation/current/reasoner-tutorial.html}.

\subsubsection{Arachne}

\textbf{Summary:} Arachne \cite{balhoff2018arachne} is an RDF rule engine written in Scala with support for efficient reasoning over large OWL RL terminologies. It implements the Rete/UL algorithm of Robert B. Doorenbos \cite{doorenbos1995production}.

\textbf{Download and Repository:} The source code is available on Github\footnote{https://github.com/balhoff/arachne}. Latest release 1.2.1 and is from January 14, 2020. The release file contains a couple of Jar files as well as a bat-file for Windows and a binary for Linux systems.

\textbf{Installation and Usage:} Initially there was little information available about installation and usage. I created an issue to ask for help and get more information\footnote{https://github.com/balhoff/arachne/issues/146}. The author responded very quickly and added sufficient information in the usage section of the README file. There is also a Protégé plugin available\footnote{https://github.com/balhoff/arachne-protege}, but it seems to be still unstable\footnote{https://github.com/balhoff/arachne/issues/5}.

\subsubsection{BaseVISor}

\textbf{Summary:} BaseVISor \cite{matheus2006basevisor} is a forward-chaining inference machine written in Java and optimized for ontological and rule-based reasoning.

\textbf{Download and Repository:} BaseVISor can be downloaded from the following project page\footnote{https://vistology.com/products/basevisor/}. The current license allows use free-of-charge for academic purposes only. For all other purposes a commercial license is to be acquired. To start the download you need to enter your name and email. You will receive a license key later on, which is required to use the software.

\textbf{Installation and Usage:} The tested version was 2.0.2 and was downloaded as ZIP archive. The archive contains some script and jar files as well as API documentation and usage information. To see if it is usable, execute test\_drive.bat (Windows) or test\_drive.sh (Linux) in the terminal (requires Java 1.5). A local test showed that BaseVISor is usable on Ubuntu 20.04, using OpenJDK 11.0.8.

\subsubsection{BORN}

\textbf{Summary:} BORN \cite{ceylan2015bayesian} is a Bayesian ontology reasoner and written in Java. Bayesian ontology language is a family of probabilistic ontology languages that allow modelling of probabilistic information about axioms in an ontology. BORN is able to work with the Bayesian ontology language BEL.

\textbf{Download and Repository:} The source code of BORN is available on Github\footnote{https://github.com/julianmendez/born}. It is provided as stand-alone software and Protégé plugin. Version 0.3.0 has been reviewed and was released in April 2017\footnote{https://sourceforge.net/projects/latitude/files/born/0.3.0/born-0.3.0.zip/download}. There are development activities from time to time. Latest commit was January 2022\footnote{https://github.com/julianmendez/born/commit/028bdf00f25594d9f1edb22ce91119d3093386b3}.

\textbf{Installation and Usage:} The Protégé plugin must be installed manually\footnote{https://github.com/julianmendez/born\#usage}. To use the standalone version, execute born.jar in the terminal. The file is located in born-standalone/target directory (see zip archive). Further usage information can be found in the Github repository as well as on a separate project page\footnote{https://julianmendez.github.io/born/}.

\subsubsection{CEL}

\textbf{Summary:} CEL \cite{mendez2009cel} is a lightweight Description Logic reasoner for large-scale biomedical ontologies in OWL 2 EL. It is written in Java and its source code is released as open source.

\textbf{Download and Repository:} The latest version 0.6.0 can be downloaded from Sourceforge\footnote{https://sourceforge.net/projects/latitude/files/cel/} and Github\footnote{https://github.com/julianmendez/cel}. There is also a Protégé plugin available, which can be downloaded from Sourceforge\footnote{https://sourceforge.net/projects/latitude/files/cel/0.6.0/de.tu-dresden.inf.lat.cel.Jar-0.6.0.zip/download}. There are sporadic development activities in the Github repository, latest commit was in January 2022\footnote{https://github.com/julianmendez/cel/commit/36b3e62bd26278d79874d0fe7ee49a6f70ff2bcc}.

\textbf{Installation and Usage:} CEL has been developed to run only on 32-bit Linux systems according to the documentation\footnote{https://github.com/julianmendez/cel\#downloading-cel}, but the source code can be compiled for other systems if needed. There is a short usage section in the README file\footnote{https://github.com/julianmendez/cel\#using-cel}. A manual with further information is available on Sourceforge\footnote{https://sourceforge.net/projects/latitude/files/cel/cel/cel-manual.pdf}. There is also a Protégé plugin available.

\subsubsection{Clipper}

\textbf{Summary:} Clipper is a conjunctive query rewriting/answering engine for Horn-SHIQ ontologies. There was no primary publication available, but it is mentioned in a few publications, such as \cite{bourguet2014trove} and \cite{shokohinia2022method}.

\textbf{Download and Repository:} It is written in Java and the source code can be found on Github\footnote{https://github.com/ghxiao/clipper}. Further information can be found in the repository, but most of it is outdated\footnote{http://www.kr.tuwien.ac.at/research/systems/clipper/download.html}.

\textbf{Installation and Usage:} The Maven build was tested successfully and Clipper CLI was shown on the terminal. No further test has been conducted. Some usage information can be found in the README file\footnote{https://github.com/ghxiao/clipper\#usage-from-cli}. Be aware that a program called DLV seems to be required, but its project page is not longer accessible\footnote{http://www.dlvsystem.com/dlvsystem/index.php/DLV}.

\subsubsection{ElepHant}

\textbf{Summary:} ElepHant \cite{sertkaya2013elephant} is a consequence-based reasoner prototype for EL+ fragment of Description Logics. It is the successor of the previously developed reasoner called CHEETAH \cite{sertkaya2011search}. The motivation behind the development of ElepHant was to improve performance and to allow usage on embedded systems with limited memory and CPUs.

\textbf{Download and Repository:} It is written in C and the source code can be found on Github\footnote{https://github.com/sertkaya/elephant-reasoner}. A binary for 64-bit Linux systems is available for download there and needs to be executed on the terminal to show a CLI. Latest commit was in 2021, but since the latest release (0.5.9) not much has changed\footnote{https://github.com/sertkaya/elephant-reasoner/compare/v.0.5.9...master}. It seems project receives a few updates from time to time, but is not developed any further.

\textbf{Installation and Usage:} The binary\footnote{https://github.com/sertkaya/elephant-reasoner/releases/tag/v.0.5.9} has been tested successfully on Ubuntu Linux 20.04. Some usage information are available in the README file\footnote{https://github.com/sertkaya/elephant-reasoner/blob/master/README}.

\subsubsection{ELK}

\textbf{Summary:} ELK \cite{kazakov2012elk} is an OWL 2 EL reasoner and written in Java. OWL 2 EL is a subset of OWL 2 that recommended when defining a large amount of classes/properties. According to the developers, ELK provides better performance compared to other reasoners, because it utilizes parallelization.

\textbf{Download and Repository:} The documentation of ELK is very extensive compared to other reasoner projects. See the project wiki on Github for installation and usage information\footnote{https://github.com/liveontologies/elk-reasoner/wiki}. ELK version 0.4.3\footnote{https://github.com/liveontologies/elk-reasoner/releases} can be downloaded as binary or pure source code. Version 0.5.0 is available as a Protégé plugin\footnote{https://oss.sonatype.org/service/local/artifact/maven/content?r=snapshots\&g=org.semanticweb.elk\&a=elk-distribution-protege\&e=zip\&v=LATEST}.

\textbf{Installation and Usage:} The Protégé plugin is one of the reasoners provided with Protégé per default. In Protégé 5.6.1 you can test ELK in version 0.4.3 and 0.5.0. ELK can also be used via the OWL API and through a command line client\footnote{https://github.com/liveontologies/elk-reasoner/wiki/GettingElk\#obtaining-elk}. Each package appears to provide different functionality.

\subsubsection{Expressive Reasoning Graph Store}

\textbf{Summary:} ERGS stands for "Expressive Reasoning Graph Store"\cite{neelam2022ergs} and is a graph store, based on JanusGraph\footnote{https://janusgraph.org/}. The project is only 3 years old and the system supports RDFS reasoning and some OWL constructs. It is written in Java.

\textbf{Download and Repository:} Source code is available on Github\footnote{https://github.com/IBM/expressive-reasoning-graph-store}, but there are no binaries available for download. However, the repository contains a file called "Docker-compose.yml", which can be used to create a Docker setup to run ERGS.

\textbf{Installation and Usage:} Information about Maven builds or how to create an ERGS repository can be found in the repository. The "docker-compose.yml" mentioned above can also be used to build a web application that provides access to ERGS\footnote{https://github.com/IBM/expressive-reasoning-graph-store\#Docker-build}.

\subsubsection{EYE and EYE JS}

\textbf{Summary:} EYE \cite{verborgh2015drawing,de2015eye} is a reasoner that allows forward and backward chaining along Euler paths. It is written in Prolog, therefore it needs a Prolog Virtual machine to run, using an Euler Abstract Machine as a core. This core is compatible with two of the most widely used Prolog engines called YAP and SWI-Prolog. The advantages of this approach is high portability. EYE has a sister project with the name "EYE JS", which aims to provide the EYE reasoner in the browser by using Webassembly\footnote{https://github.com/eyereasoner/eye-js}.

\textbf{Download and Repository:} There is an outdated website for EYE available\footnote{https://eulersharp.sourceforge.net/}, but it is superseded by another website hosted on Github\footnote{https://eyereasoner.github.io/eye/}. The source code is available on the Github repository\footnote{https://github.com/eyereasoner/eye}. The latest version downloadable version is 3.24.1. The developers are very active. At the time of checking the repository, the latest commit was a few minutes old.

\textbf{Installation and Usage:} See https://n3.restdesc.org/ for information about Semantic Web Reasoning with N3. It uses an EYE variant that is accessible via a browser. EYE also runs on your local machine. Some information about setup and usage can be found in the README file\footnote{https://github.com/eyereasoner/eye\#installation}. There are usage examples available, have a look at the "documentation" folder\footnote{Eye command line arguments and flags: \\ \url{https://github.com/eyereasoner/eye/blob/master/documentation/command\_line.md}}.

\subsubsection{FaCT++}

\textbf{Summary:} FaCT++\cite{tsarkov2006fact++} is the successor to the FaCT (Fast Classification of Terminologies) reasoner. It implements a tableaux decision procedure for the well known SHOIQ Description Logic, with additional support for data types, including strings and integers according to the developers \cite{tsarkov2006fact++}.

\textbf{Download and Repository:} The latest version 1.6.5 is available for download from the Bitbucket repository\footnote{https://bitbucket.org/dtsarkov/factplusplus/downloads/}. The last commit was in December 2017\footnote{https://bitbucket.org/dtsarkov/factplusplus/commits/650a50ce78a2f9c7fd609a2ee82b72b6e25f34ee}.

\textbf{Installation and Usage:} FaCT++ is part of the standard Protégé 5.6.1 installation, but is also available for download\footnote{https://bitbucket.org/dtsarkov/factplusplus/downloads/uk.ac.manchester.cs.owl.factplusplus-P5.x-v1.6.5.Jar}. In some cases the plugin installation fails on Windows, but there is a workaround available\footnote{\url{https://github.com/emmo-repo/EMMO/blob/7a4b825e310c5362f4c407b5c38192eb3013e37f/doc/installing\_factplusplus.md}}. It is considered usable, because of its availability in latest Protégé 5.6.1 and it is used successfully in \cite{lam2023performance}.

\subsubsection{fuzzyDL}

\textbf{Summary:} fuzzyDL \cite{bobillo2016fuzzy} is a Description Logic reasoner with support for fuzzy logic and fuzzy rough set reasoning. A quote from the project page: "\textit{It provides a reasoner for fuzzy SHIF with concrete fuzzy concepts (ALC extended with transitive roles, a role hierarchy, inverse, reflexive, symmetric roles, functional roles, and explicit definition of fuzzy sets)}"\footnote{Source: https://www.umbertostraccia.it/cs/software/fuzzyDL/fuzzyDL.html}. Although, it uses a reasoner, it seems that the system itself is a reasoner.

\textbf{Download and Repository:} There was no source code repository available, but there are two websites with further information. The main page contains information about the software itself, but also links to downloads, latest changes etc\footnote{https://www.umbertostraccia.it/cs/software/fuzzyDL/fuzzyDL.html}. The other website provides information about the Protégé plugin\footnote{https://www.umbertostraccia.it/cs/software/FuzzyOWL/index.html}. Currently version 2.3 is available for download (January 9, 2019).

\textbf{Installation and Usage:} The plugin is only available for the outdated Protégé 4.3\footnote{https://www.umbertostraccia.it/cs/software/fuzzyDL/download.html}. Use the standard procedure to install the plugin. No further installation or usage information could be found. I tried to test the plugin using Protégé 4.3, but Protégé didn't start properly\footnote{Running "./run.sh" in the terminal showed the following errors: org.protege.common.jar (org.osgi.framework.BundleException: Unresolved constraint in bundle org.protege.common [1]: \\
    Unable to resolve 1.0: missing requirement [1.0] osgi.wiring.package; (\&(osgi.wiring.package=org.w3c.dom)(version>=0.0.0)))
    org.osgi.framework.BundleException: \\
    Unresolved constraint in bundle org.protege.common [1]: \\
    Unable to resolve 1.0: missing requirement [1.0] osgi.wiring.package; (\&(osgi.wiring.package=org.w3c.dom)(version>=0.0.0)) [...]}.

\subsubsection{HermiT}

\textbf{Summary:} HermiT \cite{glimm2014hermit} is an OWL 2 reasoner, written in Java. It uses hypertableu calculus and provides support for entailment checking and also reasoning services such as class and property classification or answering SPARQL queries.

\textbf{Download and Repository:} The former project website is no longer available\footnote{http://www.hermit-reasoner.com/}. Several repositories were found on Github, but it seems that https://github.com/phillord/hermit-reasoner contains the source code of the latest version (1.3.8). Latest commit is over 6 years old\footnote{https://github.com/phillord/hermit-reasoner/commit/37ec30aced32ac81ebecc5e33fad255ddefcb4c3}. Repository has been abandoned\footnote{https://github.com/phillord/hermit-reasoner/pull/3\#issuecomment-1639673663}. There are other forks, but they seen to contain few, if any adaptions. There was no binary available on Github, but \url{https://Jar-download.com/?search\_box=HermiT} provides an unofficial one. HermiT is also available as a Protégé plugin\footnote{https://protegewiki.stanford.edu/wiki/HermiT}.

\textbf{Installation and Usage:} HermiT 1.4.3.456 is part of the standard Protégé 5.6.1 package. Interesting observation, the Protégé plugin website only contains information for version 1.3.8 and earlier\footnote{https://protegewiki.stanford.edu/wiki/HermiT}. No installation or usage documentation was available. HermiT is considered usable, because its available in the latest Protégé 5.6.1 and it is used successfully in \cite{lam2023performance}.

\subsubsection{jcel}

\textbf{Summary:} jcel \cite{mendez2012jcel} is a Description Logic EL+ reasoner hat implements a subset of OWL 2 EL. It is written in Java and uses a rule-based completion algorithm internally. It can be used as a Java library or as a Protégé plugin. Julian Mendez developed jcel, but was also part of the BORN reasoner development team.

\textbf{Download and Repository:} The source code is available on Github\footnote{https://github.com/julianmendez/jcel}. A zip archive containing the relevant files is available for download on Sourceforge\footnote{https://sourceforge.net/projects/jcel/files/jcel/0.24.1/zip/jcel-0.24.1.zip/download}. The Protégé plugin can be downloaded from the Github repository\footnote{https://github.com/julianmendez/jcel/releases}. The latest version 0.24.1 was in 2016, but latest commit was in 2022\footnote{https://github.com/julianmendez/jcel/commit/cdfb5f77312f84a6b81531d7be9974783756ff12}.

\textbf{Installation and Usage:} The jcel plugin is part of the standard Protégé 5.6.1 package. There is no dedicated page for the plugin. It is also available as a Java library, but it has to be built with Maven before it can be used. Good documentation is available\footnote{https://github.com/julianmendez/jcel\#source-code}. The library was built locally and running "jcel-standalone/target/jcel.Jar" in the terminal produced a CLI.

\subsubsection{JFact}

\textbf{Summary:} JFact\footnote{https://jfact.sourceforge.net/} is a Java port of FaCT++, an OWL DL reasoner for OWL API 3.x and 4.x. There is no dedicated publication for JFact available, but it is at least mentioned in \cite{mrozek2018aabox}.

\textbf{Download and Repository:} The project website contains some information and also points to a Github repository\footnote{https://github.com/owlcs/jfact}. On Sourceforge several versions (Jar file and Protégé plugin) until version 4.0.0 are available for download\footnote{https://sourceforge.net/projects/jfact/files/}. On Github the source code until version 4.0.2 is available to download\footnote{https://github.com/owlcs/jfact/tags}.

\textbf{Installation and Usage:} Protégé plugin 4.0.4 has been reported to work at least partially with Protégé 5.2\footnote{https://github.com/owlcs/jfact/issues/18}. It is not clear if these problems have been. JFact seems to work with (obsolete) Protégé 4.x just fine. JFact can also be used in a Java project.

\subsubsection{KAON2}

\textbf{Summary:} KAON2\cite{motik2006kaon2} provides reasoning (e.g. satisfiability decision, subsumption hierarchy computation) for SHIQ knowledge bases. It is free of charge for academic purposes only, otherwise a commercial license is required. Ontoprise GmbH, the company behind the software, filed for bankruptcy in 2012\footnote{\url{https://en.wikipedia.org/wiki/Ontoprise\_GmbH}}.

\textbf{Download and Repository:} The project website provides extensive information about KAON2\footnote{http://kaon2.semanticweb.org/}. The latest downloadable version is from 2008\footnote{http://kaon2.semanticweb.org/release/kaon2-bin-2008-06-29.zip}. According to the feedback from Boris Motik (via email), the project is no longer being developed.

\textbf{Installation and Usage:} KAON2 was developed for Java 1.5, which was released in 2004, but it can still be run on Ubuntu 20.04 using OpenJDK 11. Executing the jar file in the terminal produces a user interface. No further testing has been done.

\subsubsection{Konclude}

\textbf{Summary:} Konklude \cite{steigmiller2014konclude} is a SROIQV(D) Description Logic reasoner. According to the authors it is optimized for high throughput. It is written in C++ and uses the Qt framework\footnote{https://contribute.qt-project.org/}.

\textbf{Download and Repository:} The source code is publicly available and can be downloaded from the project website\footnote{https://www.derivo.de/en/products/konclude/download/} and on Github\footnote{https://github.com/konclude/Konclude/releases}. The latest version 0.7.0 is from 2021, so the project is considered actively maintained.

\textbf{Installation and Usage:} There are binaries for Windows, Linux and Mac OS systems that can be used without installation. A Docker container setup is also available. A local test has been conducted using the Linux binary\footnote{Konclude-v0.7.0-1138-Linux-x64-GCC-Static-Qt5.12.10.zip}. To run it, the archive was extracted and the file "Konklude.sh" executed on the terminal. It spawns a CLI with further usage information. Further usage information can be found in the README file\footnote{https://github.com/konclude/Konclude/tree/master\#usage}.

\subsubsection{LiFR}

\textbf{Summary:} LiFR \cite{tsatsou2014lifr} stands for "Lighweight Fuzzy semantic Reasoner" and is a fuzzy Description Logic reasoner written in Java. It is an extension of the Pocket KRHyper reasoner. They have extended the Description Logic interface to support additional semantics and the transformation of fuzzy operators into the native first-order clause implementation\footnote{https://github.com/lifr-reasoner/lifr\#description}. The software provides inference services such as consistency checking and fuzzy entailment.

\textbf{Download and Repository:} The source code is available on Github\footnote{https://github.com/lifr-reasoner/lifr}. There are currently no binaries available. Based on the feedback from Dorothea Tsatsou (project owner) the repository is monitored for new issues and is still maintained. Latest commit was in 2023.

\textbf{Installation and Usage:} An issue was created to ask for help, because the documentation provided was insufficient\footnote{https://github.com/lifr-reasoner/lifr/issues/2}. Dorothea Tsatsou replied very quickly and provided helpful answers. LiFR can only be used in a Java project, there is no binary or Jar file available. Later on she added further information, such as in the usage section\footnote{https://github.com/lifr-reasoner/lifr\#usage}. LiFR is considered usable based on the provided information in the Github repository.

\subsubsection{LiRoT}

\textbf{Summary:} LiRoT \cite{bento2022lirot} is a reasoner that provides reasoning for a subset of OWL 2 RL and RDF-S entailment list. It is written in C and uses the RETE algorithm. It has been optimized to use limited memory and CPUs more efficiently, making it suitable for embedded systems, such as Arduino boards. The project is just over one year old (started in 2022), which is very young in comparison to other reasoners.

\textbf{Download and Repository:} Source code is available under the terms of the CeCILL-C license\footnote{\url{http://www.cecill.info/licences/Licence\_CeCILL\_V2.1-en.html}} on a Gitlab repository\footnote{https://gitlab.com/coswot/lirot}. No binaries were found.

\textbf{Installation and Usage:} To use, first check out the Git repository and then start compiling. Afterwards, LiRot can be used inside a C project. A demo is also provided, which was successfully run on Ubuntu 20.04. See the README file for further usage information\footnote{https://gitlab.com/coswot/lirot\#how-to-use-lirot-as-a-library}. LiRot runs on Linux and Arduino systems, according to the documentation.

\subsubsection{NORA}

NORA is a scalable OWL Reasoner, which is based on NoSQL databases and Apache Spark, according to the developer\footnote{https://github.com/benhid/nora}. No publications could be found, however the following reasons speak for a mention: First of all, it is very young, the latest commit is from January 9, 2023\footnote{https://github.com/benhid/nora/commit/8d591656cb38068422fd0889d6fb8d7ca4835f9f}. A cursory look at the source code leads me to believe that it provides a relevant OWL reasoner. Also, the developer works at University Málaga\footnote{https://itis.uma.es/en/personal/antonio-benitez-hidalgo-2/} and researches Knowledge Graphs, large-scale data processing and Big Data management. No own local testing was conducted, as a working Apache Spark cluster is required, which in turn depends on a specific use case and data.

\subsubsection{Ontop and Quest}

\textbf{Summary:} Quest \cite{quest2012obda} is a reasoner and part of the Ontop\footnote{https://ontop-vkg.org/} framework, an ontology-based data access (OBDA) system. OBDA basically means that data is often read from a legacy data source, such as an RDBMS and linked to a vocabulary from an ontology. This allows virtual knowledge graphs to be created. According to the developers, Quest supports SPARQL query answering for OWL 2 QL and RDFS entailment regimes \cite{rodriguez2012quest}. Behind the scenes, SPARQL queries are converted to highly optimized SQL queries and executed in an RDBMS.

\textbf{Download and Repository:} Ontop is written in Java and the source code is available on Github\footnote{https://github.com/ontop/ontop}. The latest version of Ontop is 5.0.2 and can be downloaded from the Github repository and other sources\footnote{https://ontop-vkg.org/guide/getting-started.html}. The project is actively as the latest commit was 17 hours ago, last time I checked (June 2023).

\textbf{Installation and Usage:} Ontop has extensive documentation at https://ontop-vkg.org/guide. There is a Protégé plugin available, but it has to be installed manually. Also, Ontop can be used via a CLI\footnote{https://ontop-vkg.org/guide/cli.html} and SPARQL endpoint\footnote{https://ontop-vkg.org/tutorial/endpoint/}. The file "ontop-cli-5.1.0-SNAPSHOT.zip" has been downloaded and extracted to test Ontop. Running the file "./ontop" in the terminal produced a CLI with further information.

\subsubsection{Openllet}

\textbf{Summary:} Openllet is an OWL 2 reasoner and built on top of Pellet. It is written in Java and provides functionality to check the consistency of ontologies, compute the classification hierarchy, explain inferences, and answer SPARQL queries\footnote{https://github.com/Galigator/openllet\#openllet-is-an-owl-2-dl-reasoner}. No specific publications about Openlett could be found, but it is mentioned in \cite{singh2020owl2bench}.

\textbf{Download and Repository:} Openllet is open source and its source code as well as pre-generated Jar files can be downloaded from Github\footnote{https://github.com/Galigator/openllet}. The latest version 2.6.5 is from September 27, 2019\footnote{https://github.com/Galigator/openllet/releases}. The project is considered maintained as it receives updates from time to time (latest commit was in May 2023\footnote{https://github.com/Galigator/openllet/commit/3abccbfc0eec54233590cd4149055b78351e374d}).

\textbf{Installation and Usage:} Openllet can be used via a Jar file in your own Java project. Examples are available in the repository\footnote{https://github.com/Galigator/openllet/tree/integration/examples/src/main/java/openllet/examples}. There seems to be a Protégé plugin based on various sources\footnote{https://mvnrepository.com/artifact/com.github.galigator.openllet/openllet-protege and https://Jar-download.com/artifacts/com.github.galigator.openllet/openllet-protege/2.6.5}, but a usable plugin could not be found.

\subsubsection{OWL-RL}

\textbf{Summary:} OWL-RL is a reasoner, written in Python, for the OWL 2 RL profile\footnote{\url{https://www.w3.org/TR/owl2-profiles/\#Reasoning\_in\_OWL\_2\_RL\_and\_RDF\_Graphs\_using\_Rules}}. It is based on RDFLib\footnote{https://github.com/RDFLib/rdflib}, an RDF library that provides tools and services for working with RDF data (Parser, SPARQL and store implementations).

\textbf{Download and Repository:} The project repository is on Github\footnote{https://github.com/RDFLib/OWL-RL}. The latest version 5.2.3 (from September 13, 2021) can be downloaded there.

\textbf{Installation and Usage:} No local testing has been done, as you need to include the library in your own Python project. The README file contains some usage information\footnote{https://github.com/RDFLib/OWL-RL/blob/master/README.rst}, but overall documentation is very scarce\footnote{https://owl-rl.readthedocs.io/en/latest/installation.html}. However, the project is considered usable and maintained as the latest commit is only 2 years old and its developers are very active in other RDF projects. Therefore future refinements and developments are very likely.

\subsubsection{Pellet}

\textbf{Summary:} Pellet \cite{parsia2004pellet} is an OWL 2 DL reasoner and, according to the authors, was the first sound and complete OWL 2 DL reasoner with extensive support for reasoning with individuals, user-defined data types, and debugging support for ontologies \cite{sirin2007pellet}. It is written in Java and its source code is publicly available.

\textbf{Download and Repository:} It official website was difficult to find. In publications \cite{parsia2004pellet} and \cite{sirin2007pellet} the website http://www.mindswap.org/2003/pellet/ is mentioned, but it is not available anymore. There are a few repositories on Github which claim to provide the source code of Pellet. The repository https://github.com/severin-lemaignan/pellet seems to contain the original source code. Another repository is maintained by Stardog Union\footnote{https://github.com/stardog-union} and it contains a more recent version (v2). In the projects README file there is a note about a version 3 of Pellet\footnote{https://github.com/stardog-union/pellet/blob/master/README.md} which is no longer open source. Both repositories are considered not maintained anymore. In "severin-lemaignan/pellet" the latest commit was 2011\footnote{https://github.com/severin-lemaignan/pellet/commit/7710fb0f258c081d5dfe8f8af0f829e1a73d2250} while in "stardog-union/pellet" the latest commit was 2017\footnote{https://github.com/stardog-union/pellet/commit/4c7d16bd1811ec04117fa4cd96ed592c6cfa956b}. There are 5 open pull requests\footnote{A pull request is a contribution by the community, e.g. a patch} with no comments\footnote{https://github.com/stardog-union/pellet/pulls}.

\textbf{Installation and Usage:} The Pellet plugin is part of the standard Protégé 5.6.1 package and is still usable. It can be used via a Jar file within a custom Java project. A local test of the CLI failed, because running "pellet.sh" on Ubuntu 20.04 aborted due to compilation errors\footnote{In the following a snippet of the long error message: WARNING: An illegal reflective access operation has occurred\\
    WARNING: Illegal reflective access by com.google.inject.internal.cglib.core.$ReflectUtils$1 (file:/usr/share/maven/lib/guice.jar) to method java.lang.ClassLoader.defineClass([...])\\
    WARNING: Please consider reporting this to the maintainers of com.google.inject.internal.cglib.core.$ReflectUtils$1 \\
    WARNING: Use --illegal-access=warn to enable warnings of further illegal reflective access operations \\
    WARNING: All illegal access operations will be denied in a future release [ERROR] COMPILATION ERROR : package javax.xml.bind does not exist [...]}.

\subsubsection{RDFox}

\textbf{Summary:} RDFox \cite{nenov2015rdfox} is an RDF store with support for materialisation-based parallel Datalog reasoning (OWL 2 RL), SWRL reasoning and SPARQL query answering. According to the developers, it can efficiently process billions of triples. They provide a detailed overview of the reasoning capabilities\footnote{https://docs.oxfordsemantic.tech/reasoning.html}. A license key is required to use the system. Information about conditions or prices could not be found.

\textbf{Download and Repository:} RDFox can be downloaded for Windows, Linux and Mac OS from https://www.oxfordsemantic.tech/downloads. To receive the required license key submit the form from https://www.oxfordsemantic.tech/tryrdfoxforfree. I received a 30-day evaluation key by email. A repository containing the source could not be found.

\textbf{Installation and Usage:} RDFox 6.2 was downloaded and extracted as a ZIP archive. It contains a file that spawns a CLI when executed in a terminal. To test its usability I followed the steps in the "Getting Started" section of the documentation\footnote{https://docs.oxfordsemantic.tech/getting-started.html\#getting-started}. After starting the store, sample data was uploaded and a test query was successfully submitted. In comparison to other reasoners, RDFox is one of the easiest to setup and use.

\subsubsection{RDFSharp.Semantics}

\textbf{Summary:} RDFSharp.Semantics is an API and provides a SWRL-Reasoner with forward-chaining inference capabilities. It is an extension of the RDFSharp-API\footnote{https://github.com/mdesalvo/RDFSharp}, which provides functions to work with RDF data in general.

\textbf{Download and Repository:} It is written in C\#\footnote{\url{https://en.wikipedia.org/wiki/C\_Sharp\_(programming\_language)}} and available as open source on Github\footnote{https://github.com/mdesalvo/RDFSharp.Semantics}. The latest version 3.5.0 can be downloaded from Github and nuget\footnote{https://www.nuget.org/packages/RDFSharp.Semantics}.

\textbf{Installation and Usage:} RDFSharp.Semantics can only be used inside a custom .NET project. There is a guide with more information about installation and usage\footnote{See PDF file in latest release at the bottom, for instance:\\ https://github.com/mdesalvo/RDFSharp.Semantics/releases/tag/v3.5.0}. I have not conducted any further testing due to lack of knowledge of .NET. However, based on the development on Github and recent commits (from 2023) the project is considered usable and maintained.

\subsubsection{reasonable}

\textbf{Summary:} reasonable is an OWL 2 RL reasoner, written in Rust and available as open source on Github\footnote{https://github.com/gtfierro/reasonable}. According to the developers, it is much faster than the OWLRL and Allegro reasoners\footnote{https://github.com/gtfierro/reasonable\#performance}. There is a section about supported OWL 2 rules in the README file\footnote{https://github.com/gtfierro/reasonable\#owl-2-rules}.

\textbf{Download and Repository:} The reasoner can be installed using Cargo (Rust) or pip (Python). Docker containers are also available\footnote{\url{https://github.com/gtfierro?tab=packages\&repo\_name=reasonable}}. The developer is very active and responds promptly to issues. Besides, a static build is also available \footnote{Use file "reasonable-static
    " from https://github.com/gtfierro/reasonable/releases/tag/nightly}.

\textbf{Installation and Usage:} Use Cargo or pip to install the reasoner\footnote{https://docs.rs/reasonable/latest/reasonable/} on a local machine. See the README file for more information on Python Bindings\footnote{https://github.com/gtfierro/reasonable\#python}. At first, running the binary gave errors, so I created an issue to tell the developer about it\footnote{https://github.com/gtfierro/reasonable/issues/24}. He was very helpful and provided a fixed static build binary that ran successfully in the terminal (show a CLI).

\subsubsection{Sequoia}

\textbf{Summary:} Sequoia \cite{bate2016sequoia} is an OWL 2 DL reasoner, written in Scala and released under the terms of the GPL 3. A Scala environment is required to run Sequoia.

\textbf{Download and Repository:} The project repository is available on Github\footnote{https://github.com/andrewdbate/Sequoia}. The documentation is very sparse and it seems that the whole repository has to be downloaded before any of the Sequoia parts can be used.

\textbf{Installation and Usage:} I have not conducted ay tests due to my lack of knowledge of Scala. It seems to be maintained and usable as the latest commit was in 2020 and there are no issues in the bug tracker that indicate installation or usage problems. An issue has been opened to ask the developers to provide help with installation and a small usage example, but no response yet\footnote{https://github.com/andrewdbate/Sequoia/issues/3}.

\subsubsection{TRILL, TRILLP, TORNADO}

\textbf{Summary:} TRILL \cite{zese2013trill} is a Description Logics reasoner, written in Prolog. It supports query answering for SHOIN (D) knowledge bases. $ TRILL^{P} $ and TORNADO are based on the source code of TRILL, but are not of interest.

\textbf{Download and Repository:} The source code can be found on Github\footnote{https://github.com/rzese/trill}. The latest version 6.0.5 is from November 2022 and can be downloaded there. It is also possible to use a web based SWI-Prolog environment, which is provided by the linked Docker image\footnote{https://github.com/friguzzi/trill-on-swish}.

\textbf{Installation and Usage:} TRILL is a SWI-Prolog pack and therefore requires SWI-Prolog\footnote{https://www.swi-prolog.org/} to run. There is also a way to try out TRILL in a browser by using SWISH\footnote{https://trill-sw.eu/p/thrt.pl}. The team provides a PDF manual for TRILL on Github\footnote{https://github.com/rzese/trill/tree/master/doc}.

\subsubsection{Vampire}

\textbf{Summary:} Vampire \cite{tsarkov2004vampire, kovacs2013vampirefol} is a theorem prover written in C++, but also supports OWL DL reasoning.

\textbf{Download and Repository:} The source code is available on Github\footnote{https://github.com/vprover/vampire}. Latest version 4.7 is dated August 2022 and is available as Linux binary\footnote{https://github.com/vprover/vampire/releases}. To use it on non-Linux system, the source code has to be compiled manually\footnote{https://vprover.github.io/download.html}.

\textbf{Installation and Usage:} The downloaded Linux binary was ready to use. The following page summarizes first steps of use\footnote{https://vprover.github.io/usage.html}.

\subsubsection{VLog}

\textbf{Summary:} VLog \cite{carral2019vlog} is a Datalog-engine and rule-based reasoner (e.g. with existential rules). It is resource-saving and can handle multiple data sources, according to the developers\footnote{https://github.com/karmaresearch/vlog/wiki}.

\textbf{Download and Repository:} The source code and extensive documentation are available on Github\footnote{https://github.com/karmaresearch/vlog}. The latest version 1.3.7 (from December 2022) can be downloaded there\footnote{https://github.com/karmaresearch/vlog/releases}.

\textbf{Installation and Usage:} One way to run VLog is to compile it yourself or to use the Docker container\footnote{https://github.com/karmaresearch/vlog\#Docker}. I had to install some packages (cmake, g++, zlib1g-dev) before the compilation worked. After compiling, the file "build/vlog" was available. After executing it in a terminal, it showed a basic CLI. The wiki contains more information, such as usage examples\footnote{https://github.com/karmaresearch/vlog/wiki/Usage}.

\subsubsection{Whelk}

\textbf{Summary:} Whelk is an ELK-based reasoner, but written in Scala. It provides a Scala API and interface for OWL API. There is an integration with OBO\footnote{Open Biomedical Ontologies} tool Robot available\footnote{https://robot.obolibrary.org/reason} as well as a Protégé plugin\footnote{https://github.com/balhoff/whelk/releases/tag/v1.1.1}.

\textbf{Download and Repository:} The project repository is located on Github\footnote{https://github.com/balhoff/whelk}. The latest version 1.1.2 is from November 2022 and can be downloaded there\footnote{https://github.com/balhoff/whelk/releases}. The developer notes, that API changes are likely because the project is still under heavy development.

\textbf{Installation and Usage:} An issue was created due to problems with installation and first steps\footnote{https://github.com/balhoff/whelk/issues/217\#issuecomment-1609670170}. The developer responded quickly and provided a detailed answer. The reasoner can also be tried in a browser on the website https://balhoff.github.io/whelk-web/. Provided example was tested successfully.

\subsection{Systems using a third-party reasoner}

In the following is a list of all systems, which use a foreign standalone reasoner. During the research it was not always clear whether a system was a standalone reasoner or not. This distinction is important to some extent, because one might want to find out capabilities/functionality of a reasoner, for instance. They are also listed separately to support future research. Only a few systems have been tested in detail.

\subsubsection{AberOWL}

AberOWL \cite{slater2016using} is a framework for ontology-based access to biological data. It consists of an ontology repository and some web services for reasoning. It provides OWL 2 EL reasoning using the ELK reasoner. The project site is http://aber-owl.net/.

\subsubsection{BUNDLE}

BUNDLE \cite{riguzzi2013bundle} is written in Java and provides reasoning for probabilistic ontologies using the FaCT++ reasoner (and others). It uses the DISPONTE\footnote{https://ml.unife.it/disponte/} approach, i.e. each axiom is accompanied by a probability number (e.g. 0.6). This number represents the probability that a certain axiom is considered true (epistemic probability). The project page https://ml.unife.it/bundle/ contains numerous downloads as well as documentation. The latest commits were made in 2023. BUNDLE can be tried out in the browser. The following website contains pre-built examples https://bundle.ml.unife.it/examples. The Bitbucket repository provides information on getting started and usage\footnote{https://bitbucket.org/machinelearningunife/bundle/src/master/README.md and \\https://bitbucket.org/machinelearningunife/bundle/wiki/BUNDLE\%203.0.x}.

\subsubsection{Chainsaw}

Chainsaw \cite{tsarkov2012chainsaw} is a meta reasoner that uses third-party reasoners such as ELK for a given reasoning task. It was designed to handle large ontologies. Internally, it uses the following approach: for each reasoning query it generates a module of the ontology and select the most appropriate reasoner to process it. They have a Bitbucket repository\footnote{https://bitbucket.org/ignazio1977/chainsaw/src/master/}, which contains the source code. The latest commit is from 2022. Downloads can be found on Sourceforge\footnote{https://sourceforge.net/projects/chainsaw/}.

\subsubsection{ComR}

ComR \cite{wang2019comr} is a reasoner prototype to demonstrate an approach for faster ontology classification. According to the authors, the time savings are up to 90\% compared to similar reasoners such as HermiT, FaCT++ or Pellet. To achieve this they combine an OWL 2 EL reasoner with an OWL 2 reasoner for ontology classification in SROIQ. The non-EL-part is done by a slower OWL 2 reasoner. A project repository or website could not be found, so it is considered no longer maintained.

\subsubsection{COROR}

COROR \cite{o2011coror} stands for "COmposable Rule-entailment Owl Reasoner" and is a reasoner optimized for embedded systems with limited memory and CPUs. It was developed to research composition algorithms for rule-entailment OWL reasoners. A project repository or website could not be found, so it is considered to be no longer maintained.

\subsubsection{DLEJena}

DLEJena \cite{meditskos2010dlejena} combines Apache Jena's forward-chaining rule engine and the Pellet reasoner to provide OWL 2 RL reasoning. A project repository or website could not be found, so it is considered to be no longer maintained.

\subsubsection{DRAOn}

DRAOn is written in Java and provides reasoning for networks of ontologies. It supports both standard Description Logics semantics for non-distributed reasoning and the Integrated Distributed Description Logics (IDDL) semantics for distributed reasoning \cite{le2013draon}. The OWL API and Alignment API 4.0\cite{david2011alignment} are used internally. The HermiT reasoner is used to find inconsistencies in ontologies and alignments. The official project page is no longer accessible\footnote{http://iddl.gforge.inria.fr/} and a source code repository, or something similar, could not be found.

\subsubsection{ELOG}

ELOG \cite{niepert2011log} is a prototype of an EL++ log-linear Description Logic reasoner, written in Java. Its source code is publicly available and can be found on Google Code\footnote{https://code.google.com/archive/p/elog-reasoner/}. ELOG uses foreign reasoners, such as HermiT or Pellet. There are some files available for download, while the file "elog.zip" contains files to run ELOG. After extracting the "elog.zip" archive, run the "elog.Jar" in a terminal to spawn a CLI. You can also use a dedicated web service\footnote{http://executor.informatik.uni-mannheim.de/systems/elog/}, to check out the reasoner. The last commit was in 2014\footnote{https://code.google.com/archive/p/elog-reasoner/source/default/commits}. There is no indication that the project is still maintained.

\subsubsection{Hoolet}

Hoolet\footnote{http://owl.man.ac.uk/hoolet/} provides OWL 2 DL reasoning using the theorem prover Vampire. A dedicated publication could not be found, but a short presentation is available\footnote{http://www.daml.org/meetings/2004/05/pi/pdf/Hoolet.pdf}. It is also mentioned in other publications, for instance \cite{jang2004bossam}. The downloadable file has to be extracted and provides a file called "hooletGUI", which opens an user interface. Based on my findings I consider the project to be no longer maintained.

\subsubsection{Hydrowl}

Hydrowl \cite{stoilos2014hydrowl} is a query answering system for OWL 2 DL ontologies, using foreign reasoners, such as HermiT or OWLIM. The project page http://www.image.ece.ntua.gr/\~gstoil/hydrowl/ is no longer available. There is no evidence that the project is still maintained.

\subsubsection{HyLAR}

HyLAR \cite{terdjimi2015hylar} stands for "Hybrid Location-Agnostic Reasoner" and partially supports OWL 2 RL reasoning. It is written in JavaScript and uses the reasoner of an abandoned project called JSW Toolkit\footnote{https://code.google.com/archive/p/owlreasoner/}\cite{terdjimi2016hylarplus}. The project page is hosted on Github\footnote{https://github.com/ucbl/HyLAR-Reasoner} and the latest commit was in 2021. HyLAR is also available as an NPM package\footnote{https://www.npmjs.com/package/hylar}.

\subsubsection{Minerva}

Minerva \cite{sim2006minerva,zhou2006minerva} is an ontology storage and inference system, using the Racer and Pellet reasoner. The project is mentioned in some publications, but a project page or source code repository could not be found.

\subsubsection{MORe}

MORe \cite{armas2012more} stands for "Modular OWL Reasoner" and is a system for classifying OWL 2 ontologies. It is written in Java and uses the OWL 2 reasoner HermiT and the OWL 2 EL Reasoner ELK. It seems that the reasoner is no longer usable according to one of the issues on Github\footnote{https://github.com/anaphylactic/MORe/issues/1}. The latest commit was 7 years ago, the project is considered no longer maintained.

\subsubsection{NoHR}

NoHR \cite{lopes2017nohr} stands for "Nova Hybrid Reasoner" and is a system that uses various reasoners, such as HermiT or ELK) for reasoning tasks. It is written in Java and can be used as Protégé plugin or via the Java API. NoHR was included because it provides reasoning for different OWL profiles and supports a rule engine (XSB Prolog\footnote{https://xsb.sourceforge.net/}). The project is hosted on Github\footnote{https://github.com/NoHRReasoner/NoHR}, and I was able to successfully load the plugin into Protégé, so I assume the project is still usable. The latest commit was in 2019 though.

\subsubsection{PAGOdA}

PAGOdA \cite{zhou2015pagoda} provides conjunctive query answering for OWL 2 ontologies. Internally it uses the Datalog reasoner RDFox and the OWL 2 reasoner HermiT. It can be used either using a Jar file or via the Java API. The source code is hosted on Github\footnote{https://github.com/KRR-Oxford/PAGOdA}. The latest version 2.1.2 was in 2015, but the latest commit was in February 2023. I assume the project gets small updates sporadically.

\subsubsection{PROSE}

PROSE \cite{wu2016prose} is a plugin-based paraconsistent OWL reasoner that uses a third-party reasoner, such as Pellet or HermiT. Paraconsistent OWL ontologies are characterized by the fact, that they can contain conflicting axioms. No project page or source code repository could be found.

\subsubsection{R2O2*}

$ R_{2}O_{2}* $\cite{li2020r2o2star} is a meta reasoner that selects the most appropriate reasoner, such as ELK or FaCT++, for a given reasoning task. Robustness and efficiency are used as selection criteria, for instance. It uses the OWL API 3.4. The source code and other material is hosted on Github\footnote{https://github.com/liyuanfang/r2o2-star}. I tried out the reasoner, but compilation failed due to unfulfilled dependencies\footnote{[ERROR] Failed to execute goal on project r2o2-star: Could not resolve dependencies for project edu.monash.infotech:r2o2-star:Jar:1.0: Failed to collect dependencies at io.wasiluk:weka-xgboost:Jar:0.2.0 -> biz.k11i:xgboost-predictor:Jar:0.3.0: Failed to read artifact descriptor for biz.k11i:xgboost-predictor:Jar:0.3.0: Could not transfer artifact biz.k11i:xgboost-predictor:pom:0.3.0 from/to bintray-komiya-atsushi-maven (http://dl.bintray.com/komiya-atsushi/maven): Transfer failed for http://dl.bintray.com/komiya-atsushi/maven/biz/k11i/xgboost-predictor/0.3.0/xgboost-predictor-0.3.0.pom: Unknown host dl.bintray.com:}. Latest commit was in 2019.

\subsubsection{REQUIEM}

REQUIEM is an OWL 2 QL reasoner to demonstrate a specific query rewriting algorithm. It uses HermiT reasoner for reasoning tasks. The project page https://www.cs.ox.ac.uk/isg/tools/Requiem/ mentions \cite{perez2010tractable} as the primary publication, but does not contain any reference to REQUIEM. It was mentioned in passing in \cite{wu2012oraclerequiem}. On Github there is one repository containing the source code\footnote{https://github.com/ghxiao/requiem} and another repository that provides access to REQUIEM via a Docker container\footnote{https://github.com/justin2004/pomify-REQUIEM}. The latest commit to the source code repository was in 2013.

\subsubsection{RuQAR}

RuQAR \cite{bak2014ruqar} is a query answering and reasoning framework for OWL 2 RL ontologies. It uses the HermiT reasoner for TBox reasoning. RuQAR has been mentioned in passing a few times in some publications. No project page could be found.

\subsubsection{Screech}

Screech \cite{hitzler2005screech} seems to be a prototype for OWL ABox reasoning with approximation using KAON2. This approach is useful in applications where efficient queries are mot important than correctness. The project page http://logic.aifb.uni-karlsruhe.de/screech is no longer available.

\subsubsection{TrOWL: Quill and REL}

TrOWL \cite{pan2012exploiting,thomas2010trowl} is a reasoning infrastructure for OWL 2. It uses Quill reasoner and REL, an OWL 2 EL reasoner, which is based on ELK reasoner. The official links to both reasoners are broken\footnote{https://www.w3.org/2001/sw/wiki/OWL/Implementations}. The source code of TrOWL is hosted on Github, but without documentation\footnote{https://github.com/TrOWL/core}. Fun fact: the repository links to the page http://trowl.eu/, which contains pornographic content!

\subsubsection{WSClassifier and WSReasoner}

WSClassifier \cite{song2013wsclassifier} is a reasoner for ALCHI(D) ontologies. In the reasoner list\footnote{http://owl.cs.manchester.ac.uk/tools/list-of-reasoners/} another reasoner with the same name (WSReasoner \cite{song2012wsreasoner}) is mentioned. It uses the ConDOR and HermiT reasoner internally. There is a Google Code repository for WSClassifier, but it only provides some information and a handful download links\footnote{https://code.google.com/archive/p/wsclassifier/}. Each download link leads to a Google Drive folder.

\subsection{(Probably) unusable OWL reasoners}

In this section you will find a list of all OWL reasoners that are considered (probably) not usable. Some of the reasoners have been manually tested.

\subsubsection{BOSSAM}

BOSSAM \cite{jang2004bossam} is a RETE-based inference machine that supports reasoning for OWL and SWRL ontologies, and RuleML rules. In comparison to similar reasoners, it supports negation-as-failure and classical negation. According to the authors, it handles dynamic and conflicting knowledge bases better than the competition. It is written in Java, but is closed source. The project page does not provide a working download link\footnote{https://bossam.wordpress.com/}. Latest activity was in 2007\footnote{https://bossam.wordpress.com/2007/08/18/new-api/}.

\subsubsection{CB OWL 2 Horn Reasoner}

CB \cite{kazakov2015consequence} is a reasoner for Horn SHIQ ontologies. It is written in OCaml and has been developed as part of the ConDOR\footnote{https://www.cs.ox.ac.uk/isg/projects/ConDOR/} project. They host the source code on Github\footnote{https://github.com/ykazakov/cb-reasoner} and the latest commit is 13 years old. There are no binaries to download, so compilation is required before the reasoner can be used. The reasoner should run on Windows, Linux and MacOS X, but there are problems indicating that this may no longer be the case anymore (e.g. on Ubuntu 14.04\footnote{https://github.com/ykazakov/cb-reasoner/issues/2}). The compilation\footnote{https://github.com/ykazakov/cb-reasoner/blob/master/INSTALL} failed on my local machine\footnote{I had to install ocamlbuild, which fixed one error, but I ran into the following which I couldn't solve: "/bin/sh: 1: ocamlopt: not found"}.

\subsubsection{Cereba Engine}

It was difficult to find enough information about the Cereba Engine. In \cite{mishra2011semantic, khamparia2017comprehensive} it is described as a software that provides logical reasoning through ABox and TBox for DAML+OIL\footnote{https://www.w3.org/TR/daml+oil-reference/}, OWL and ontologies. In \cite{khamparia2017comprehensive} there is a reference to a website with further information\footnote{https://www2003.org/cdrom/papers/poster/p087/Poster87.html}. A dedicated project page or source code repository could not be found, so the project is considered abandoned.

\subsubsection{CICLOP}

CICLOP is a Description Logic reasoner that supports role hierarchy and inverse roles, for instance. There is no primary publication, but it is mentioned in \cite{ciclop2003using} and \cite{ciclopensaisoptimized}. Binaries or source code for own tests could not be found.

\subsubsection{DBOWL}

DBOWL \cite{dbowl2012evaluating} is a scalable reasoner (ABox focus), optimized for ontologies with billions of triples. It provides extensive support for OWL 1 Description Logic. DBOWL uses RDBMS for ontology storage and instance classification. The official project page is https://khaos.uma.es/dbowl, but it does not contain any information. Other web sites could not be found.

\subsubsection{Deslog}

Deslog \cite{wu2012deslog} is a shared-memory parallel reasoner for Description Logic ALC. It is written in Java and uses the OWL API. It uses parallel computing to speed up reasoning, which distinguishes it from the more classic reasoners. Web sites such as the project page or a Github repository could not be found.

\subsubsection{DistEL}

DistEL \cite{distel2014developing} is a Peer-to-Peer based, distributed reasoner that can be used to classify EL+ ontologies. It is written in Java and uses Redis\footnote{https://redis.io/} for data storage. The source code is hosted on Github\footnote{https://github.com/raghavam/DistEL}. The latest commit was in 2016. The code needs to be compiled before use as there are no binaries or scripts available for download.

\subsubsection{DLP}

DLP \cite{patel1999dlp} stands for "Description Logic Programs" and is a Description Logic system that contains sound and complete reasoners for expressive Description Logics. It can be used for satisfiability checks for propositional modal logics. DLP was developed to study various optimization techniques. The official site http://www.bell-labs.com/user/pfps/dlp still exists, but only shows a 404 page not found error.

\subsubsection{DReW}

DReW \cite{xiao2010drew} is a query answering system for LDL+ ontologies and provides reasoning for Description Logic programs over LDL+ ontologies. According to the authors, it is optimized to handle large ontologies very well. It is written in Java and the source code is hosted on Github\footnote{https://github.com/ghxiao/drew}. The latest version 0.3.0 beta 3 was released in March 3, 2013 and the latest activity in the repository was in 2015. I downloaded and extracted the 0.3.0 beta 3 archive and started the CLI on Ubuntu 20.04. But first I had to complete the following steps: (1) set the global variable \textit{DREW\_HOME} to the path of the extracted ZIP folder. (2) rename the file \textit{drew-0.3-beta-3.Jar} to \textit{drew-0.3-beta-2.Jar} in the lib folder. Without the second step, an error will occur\footnote{Error: Unable to access Jarfile ...}. DReW requires DLV\footnote{http://www.dlvsystem.com/dlv/} to run (at least partly), but DLV is no longer available.

\subsubsection{ELLY}

ELLY \cite{siorpaes2010elp} is an ELP reasoner that is based on the ISIS Datalog reasoner\footnote{https://www.iris-reasoner.org/}. ELP is the decidable part of the Semantic Web Rule Language (SWRL), that allows polynomial time reasoning. By using the OWL API it supports reasoning for OWL 2 EL and OWL 2 RL. Further information about the ISIS Datalog reasoner could not be found. This website is incomplete and contains unrelated content. ELLYs project page is hosted on Sourceforge\footnote{https://elly.sourceforge.net/}.

\subsubsection{FLOwer}

$FL_0$wer \cite{michel2019flower} is a reasoner for the Description Logic $FL_0$ that provides efficient TBox reasoning with value restrictions. It is a prototype and was developed as part of a bachelor thesis\footnote{in German: https://lat.inf.tu-dresden.de/research/theses/2017/Mic-Bac-17.pdf}, is written in Java. The source code is available on Github\footnote{https://github.com/attalos/fl0wer}, but there are no binaries to download. For your own testing, you need to build the code using Maven\footnote{https://maven.apache.org/}. The latest commit was in 2019, so an issue was opened to find out if there are plans for future development\footnote{https://github.com/attalos/fl0wer/issues/1}. There has been no response yet.

\subsubsection{F-OWL}

F-OWL \cite{zou2005fowl} is an F-Logic based inference machine for RDF and OWL. It uses the XSB
logic programming system with the Flora-2 extension, which provides an F-logic frame-based representation layer. Neither a project page nor source code repository could be found, so the project is considered abandoned.

\subsubsection{HS-Reasoner}

HS-Reasoner is an OWL 2 DL reasoner written in Haskell. The project is hosted on Github\footnote{https://github.com/agnantis/hs-reasoner}. According to the developer, HS-Reasoner was developed to study possibilities of functional programming for reasoners. An issue has been created for a short introduction on how to install and use it\footnote{https://github.com/agnantis/hs-reasoner/issues/1}. I don't have any knowledge in Haskell and without any information on install and use, there was no way for me to determine if it is usable or not. The latest commit was in 2019, so it is possible that the project is still maintained.

\subsubsection{leancor}

leancor is a Description Logic reasoner and a fork of leanCoR. It is written in Prolog and the source code is hosted on Github\footnote{https://github.com/adrianomelo/leancor}. Information about installation and usage was scarce and scattered in various files. The installation failed on my local machine (Ubuntu 20.04). It seems that leancor uses the leanCoP theorem prover\footnote{http://www.leancop.de/index.html} internally\footnote{https://github.com/adrianomelo/leancor/blob/master/leancop.sh}. leancor is considered abandoned because the latest commit was in 2015 and the project page http://www.leancor.org/ is no longer accessible.

\subsubsection{LillyTab}

LillyTab \cite{wullinger2018supporting} is an OWL 2 reasoner for SHOF(D) Description Logic. It is written in Java and its source code is hosted on Github\footnote{https://github.com/pwub/lillytab}. The latest version 1.12 was released in 2015. According to the developer, LillyTab was developed as part of his thesis (\cite{wullinger2018supporting}, page 79). For this reason and the fact that latest development activity was in 2018, I assume that the project is no longer maintained.

\subsubsection{Mini-ME and Mini-ME Swift}

Mini-ME \cite{scioscia2018minime} (short for Mini Matchmaking Engine) is a reasoner for ALN (attributive language with unqualified number restrictions) Description Logic (DL). It is written in Java and optimized for use on mobile devices (Semantic Web of Things). There are two variants available:  \textbf{Tiny-ME}\footnote{http://swot.sisinflab.poliba.it/tinyme/} comes with a C-, Java- and Object-C API. \textbf{Mini-ME} Swift\footnote{http://swot.sisinflab.poliba.it/minime-swift/index.html} is an iOS port, written in Swift\footnote{Swift project page: https://developer.apple.com/swift/}. Mini-ME is only distributed for the purpose of academic evaluation and review only, no other use is allowed. On the download page http://swot.sisinflab.poliba.it/minime/\#download many links are broken. Mini-ME Swift is only available for iOS and MacOS\footnote{http://swot.sisinflab.poliba.it/minime-swift/}. No source code repository could be found. Mini-ME was tried locally. After downloading and extracting the "Mini-ME OWLLink package", the file "start-minime.sh" was executed on the terminal. It started a server which was accessible at localhost:8080, but a 404 error was displayed. Exception messages were displayed on the terminal each time I visited the index page. The library versions weren't tried, so they may still be usable. Protégé plugin is reported to work with Protégé 4.3 and 5.1\footnote{http://swot.sisinflab.poliba.it/minime/plugin.html}, but no local tests have been conducted due to missing plugin files.

\subsubsection{O-DEVICE}

O-DEVICE \cite{meditskos2008odevice} is a rule-based object oriented OWL reasoner using the CLIPS rule engine. The latest activity in the project repository was in 2013 and the latest version is from 2010. For this reason the project is considered abandoned. No local testing has been done, as most of the files are from 2009.

\subsubsection{OWLgres}

OWLgres \cite{stocker2008owlgres} is a Description Logic lite reasoner using PostgreSQL. A project page or source code repository could not be found. At https://www.semanticweb.org/wiki/Owlgres.html there is a link to the project page http://pellet.owldl.com/owlgres, but it is no longer accessible.

\subsubsection{OWLIM}

OWLIM \cite{bishop2011owlim} is a family of Semantic Web components and contains a reasoner. A project page or source code repository could not be found.

\subsubsection{Pocket KRHyper and KRHyper}

Pocket KRHyper \cite{sinner2005krhyper} is a Description Logic reasoner, highly optimized to run on embedded systems. Its predecessor is KRHyper. Own tests weren't possible due to the official download link http://www.uni-koblenz.de/$\sim$\%7B\%7Diason/downloads is not working anymore.

\subsubsection{Pronto}

PROTON \cite{papadakis2011proton} is a reasoner for managing temporal information in OWL ontologies. The project is hosted on Github and provides the source code, documentation and help scripts\footnote{https://github.com/klinovp/pronto}. The installation instructions weren't sufficient enough and own tests on Ubuntu 20.04 failed, because of a missing file\footnote{Running "pronto.sh" in the terminal resulted in the error that "dist/lib/pronto.Jar" is not available}.

\subsubsection{QueryPIE}

QueryPIE \cite{urbani2011querypie} is a reasoner that uses backward reasoning and is optimized to handle ontologies with billions of triples. The project is hosted on Github\footnote{https://github.com/jrbn/querypie}. Its worth mentioning, because it is one of the few reasoners that can handle large ontologies. The latest activity on the repository was in 2017. Downloading and building with Ant was successful, but using it requires a custom Java program and involves additional software such as Hadoop\footnote{https://hadoop.apache.org/}, so I stopped further testing.

\subsubsection{SPOR}

SPOR \cite{argyridis2015spor} stands for "SPatial Ontology Reasoner" and supports efficient reasoning on Geographic Object-Based Image Analysis (GEOBIA) ontologies using fuzzy, spatial, and multi-scale representations. The project is hosted on Github, but no binaries are provided\footnote{https://github.com/ArArgyridis/SPOR}. No installation or usage instructions are provided, so no tests have been performed. The reasoner can also be used via Gnorasi\footnote{https://github.com/gnorasi/gnorasi} according to \cite{argyridis2015spor}.

\subsubsection{QuOnto}

QuOnto \cite{lembo2013quonto} is a OWL reasoner for the OWL 2 QL profile, optimized to achieve superior performance in classifying OWL 2 QL ontologies, compared to similar reasoners. No project page or source code repositories could be found.

\subsubsection{Racer and RacerPro}

Racer is a Description Logic SRIQ(D) reasoner. It was written in Java and Common Lisp. According to the project page\footnote{https://www.ifis.uni-luebeck.de/~moeller/racer/index.html} it is the successor of RacerPro \cite{haarslev2012racerpro}. This is interesting, because publications about RacerPro are 11 years older than the release date of Racer. The source code of Racer is hosted on Github\footnote{https://github.com/ha-mo-we/Racer} and licensed under the terms of the BSD-3-clause license. The latest release was in 2014 and the latest commit was in 2018. Because of the advanced knowledge in Common Lisp skills required, no testing has been done. Although, Racer and Racer Pro may run still, but are considered unusable, e.g. because downloads are no longer available\footnote{User Guide: https://github.com/ha-mo-we/Racer/blob/master/doc/users-guide-2-0.pdf, on page 6 it links to http://www.racer-systems.com/products/download/index.phtml which is not available anymore}, for instance. The project appears to have been abandoned.

\subsubsection{Rat-OWL}

Rat-OWL \cite{giordano2017ratowl} is a reasoner for the non-monotonic extension of Description Logic. According to the publication it is only available for Protégé 4.3 (released in 2013\footnote{\url{https://protegewiki.stanford.edu/wiki/P4\_3\_Release\_Announcement}}), using OWL API 3.4. A project page or source code repository could not be found. The Protégé plugin was also no longer available\footnote{Checked \url{https://protegewiki.stanford.edu/wiki/Protege\_Plugin\_Library} and did a basic Internet search}.

\subsubsection{SHER}

SHER \cite{dolby2009sher} stands for "scalable highly expressive reasoner" and has a good performance, according to the authors. However, the project website is no longer accessible \footnote{http://www.alphaworks.ibm.com/tech/sher}.

\subsubsection{Snorocket}

Snorocket \cite{metke2013snorocket} is a highly optimized reasoner for a specific subset of the OWL 2 EL profile. The project is hosted on Github\footnote{https://github.com/aehrc/snorocket} and can be used via the OWL API or via the Protégé plugin\footnote{https://protegewiki.stanford.edu/wiki/Snorocket}. The version 3.0.0 is listed as the latest version, but no downloads could be found\footnote{\url{https://protegewiki.stanford.edu/wiki/Snorocket\_3.0.0}}. The same goes for version 2.0.0\footnote{\url{https://protegewiki.stanford.edu/wiki/Snorocket\_2.0.0}}. Although the latest commit was in 2019, the project is considered abandoned due to the lack of download links.

\subsubsection{SoftFacts}

SoftFacts \cite{straccia2010softfacts} is an information retrieval system for relational databases that provides reasoning, e.g. instance query answering. There is a project with the same name, but it is empty\footnote{https://github.com/straccia/SoftFacts}. No usable source code or binaries could be found.

\subsubsection{SparkEL}

SparkEL \cite{mutharaju2016sparkel} is a distributed OWL 2 EL reasoner using Apache Spark\footnote{http://spark.apache.org/}. It is written in Scala and the source code is hosted on Github\footnote{https://github.com/raghavam/sparkel}. The latest commit was in 2016 and the binaries provided are outdated. I tried to install it using the install script\footnote{https://github.com/raghavam/sparkel/blob/master/script/install-all.sh}, but it failed because dependencies, such as SBT 0.13.9, weren't accessible anymore. SparkEL may still be usable, but it is doubtful due to outdated dependencies.

\subsubsection{SPOWL}

SPOWL \cite{liu2017spowl} uses Apache Spark to provide reasoning for large OWL ontologies. It acts as a compiler that maps TBox axioms to Spark programs. No source code or binaries were available, so the project is considered abandoned.

\subsubsection{SWRL-IQ}

SWRL-IQ \cite{elenius2012swrliq} is a plugin for the outdated Protégé version 3. According to the authors it combined features that no other reasoning/query system had at the time, such as constraint-solving based on CLP(R) (Constraint Logic Programming with Reals) as well as SWRL extensions for non-monotonic aggregation, limited higher order logic. The plugin page\footnote{https://protegewiki.stanford.edu/wiki/SWRL-IQ} on the Protégé wiki was updated in 2012. The manual may still be useful \footnote{\url{https://protegewiki.stanford.edu/images/5/57/SWRL-IQ\_manual.pdf}}. The project page \footnote{https://www.onistt.org/display/SWRLIQ/SWRL-IQ} is no longer accessible and the project is considered abandoned.

\subsubsection{TReasoner}

TReasoner \cite{grigorev2013treasoner} is a reasoner for SHOIQ(D) that implements a tableau algorithm. The project is hosted on Google Code\footnote{https://code.google.com/archive/p/treasoner/}. The latest commit was in 2014. There is no documentation available, so I assume it is only usable within a custom Java project.

\subsubsection{WebPIE}

WebPIE \cite{urbani2010webpie} is an OWL/RDFS inference machine using Hadoop\footnote{https://hadoop.apache.org/}. The reasoning is being mapped on Map and Reduce operations, allowing distributed, more performant processing of large quantities of data. According to the project website https://www.few.vu.nl/~jui200/webpie.html\#sourcecode the project is no longer maintained. Some links in the documentation are broken and the only available file was published in 2012. Due to the outdated files and the need to setup a cluster on your own\footnote{https://www.few.vu.nl/~jui200/webpie.html\#tutorial}, no further testing have been conducted. WebPIE might still usable, but it is doubted.

\subsubsection{Wolpertinger}

Wolpertinger \cite{rudolph2017wolpertinger} is a fixed-domain OWL reasoner. OWL is based on the Open World assumption, which means, an axiom can be true regardless if it is known to be true or not. There are applications where this can lead to undesirable (because contradictory) situations. The reasoner doesn't use the standard model-theoretical semantics, instead the domain is reduced to an explicitly given set of axioms. Wolpertinger is written in Java and the source code is hosted on Github\footnote{https://github.com/wolpertinger-reasoner/Wolpertinger} (latest commit was in 2019). There are no binaries available. I tried to build one myself but ran into errors, that have been reported in an issue\footnote{https://github.com/wolpertinger-reasoner/Wolpertinger/issues/2}.

\section{Conclusion}

In this work 73 OWL reasoners and 22 systems (using a third-party reasoner) were analyzed and metadata such as usability and maintenance status was collected. All the OWL reasoners found that were considered usable and maintained, need to be analyzed in more detail and with regards to their applications. For instance, not all OWL 2 DL reasoners fully support the OWL 2 DL profile. Also, some systems were developed using outdated development environments such as Java 1.5, which was released 2011. All systems are described in a more general terms as they are currently exist.

In the following is a list of positive observations:

\begin{enumerate}
    \item \textbf{Rooted in science:} Most of the analyzed systems are mentioned in at least one publication. It is either a system description paper or an evaluation/comparison of two or more reasoners. Many papers on the underlying methods and describe them in detail, which helps to understand the inner workings of a system.
    \item \textbf{Diverse software:} OWL reasoners are developed in a variety of languages and environments. Java is the most prominent one. The functional scope of each individual system is also diverse. Each language/environment has its own advantages that make it ideal for specific use cases, such as high performance computing or embedded systems.
    \item \textbf{Constant interest}: New projects have been started in recent years, which may be an indicator that there is still sufficient research interest in this area. Many projects host their code on Github, which also encourages new people to contribute.
\end{enumerate}

The negative observations are summarized below:

\begin{enumerate}
    \item \textbf{Inadequate documentation:} Most of the projects provide little or no documentation to their users. It seems that the authors assume that their users have the necessary knowledge in any software environment they use, be it a simple Java binary, a Hadoop or Haskell/Scala setup. Also, some projects have split their documentation into different files, or even whole web sites, making it much harder to  learn about a particular system. Scientific publications may be considered as part of the documentation, but they usually do not contain any end-user information (such as installation manual or usage introductions).
    \item \textbf{Short term usage:} A large percentage of the OWL reasoners had prototype status to demonstrate a novel approach. This may partly explain why the documentation was scarce in comparison to other projects. However, prototypes are not of interest to people who want to use a stable basement for a new software, whether they are scientists or not. Poor documentation can also hinder new users who want to learn about a software.
    \item \textbf{Bad maintenance situation:} Only 28 out of 73 OWL reasoners are actively maintained, which means they have had some development activity during the last 3 years. Few had project repositories set up in a way that would support long-term development. Continuous integration pipelines or even (unit-, system-, ...) tests were scarce. Some projects on Github had more than 10 open pull requests unanswered, meaning that there were people who wanted to contribute code changes, but the author ignored them. Similar observations have been made in \cite{lam2023performance}.
\end{enumerate}

\section{Future work}

As mentioned in the previous section, all usable and maintained OWL reasoners found need to be analyzed in more detail and with regard to their applications. Verification of my findings, especially for those systems where my knowledge is very limited, would be important. In some cases I have created an issue in the project repository, which could be a good starting point for further investigation. Performance comparisons are also of interest due to the hardware developments in the recent years (e.g. parallel computing, GPU computing). Only a few OWL reasoners have used parallel computing at all.

The part of the Semantic Web/Logic community working with OWL reasoners should consolidate the current state. It is important to find out, which systems are still sufficient enough to handle current use cases. It is also important to provide the necessary documentation for end users to try them out. I am afraid we can only salvage what is left of systems like HermiT, but this could be the first step towards new developments.

For all the reasons mentioned above, I claim that a large number of analyzed OWL reasoners are not of interest to companies, especially in the areas of software development and knowledge organization. Many systems are almost black boxes to their users, with poor documentation, outdated dependencies and no support from their developers. OWL reasoners such as HermiT are still widely used, although they are being abandoned, because there are not many alternatives available. Therefore surveys/studies that further investigate this issue are very important.

\section{Acknowledgement}

This work was supported by a grant from the German Federal Ministry of Education and Research (BMBF) for the KI-Werk Projekt (https://www.cbasynergy.net/cba/ki-werk.html).

\section{Appendix}

\begin{table}[H]
    \centering
    \begin{tabular}{|l|}
        \hline
        \textbf{OWL Reasoner}                  \\ \hline
        AllegroGraph                           \\ \hline
        Arachne                                \\ \hline
        BaseVISor                              \\ \hline
        BORN                                   \\ \hline
        CEL                                    \\ \hline
        ElepHant                               \\ \hline
        ELK                                    \\ \hline
        Expressive Reasoning Graph Store (IBM) \\ \hline
        EYE                                    \\ \hline
        EYE.js                                 \\ \hline
        jcel                                   \\ \hline
        JFact                                  \\ \hline
        Konclude                               \\ \hline
        LiFR                                   \\ \hline
        LiRoT                                  \\ \hline
        Ontop (with Query reasoner)            \\ \hline
        Openllet                               \\ \hline
        OWLRL                                  \\ \hline
        RDFox                                  \\ \hline
        RDFSharp.Semantics                     \\ \hline
        reasonable                             \\ \hline
        TRILL (and TRILLP and TORNADO)         \\ \hline
        Vampire                                \\ \hline
        VLog                                   \\ \hline
        Whelk                                  \\ \hline
    \end{tabular}
    \caption{List of usable and maintained OWL reasoners \label{table-usable-maintained-owl-reasoners}}
\end{table}

\begin{table}[H]
    \scriptsize
    \begin{tabular}{|l|l|l|}
        \hline
        \textbf{Reasoner}                      & \textbf{Maintenance}     & \textbf{Is usable}                                                  \\ \hline
        AllegroGraph                           & maintained               & usable                                                              \\ \hline
        Arachne                                & maintained               & usable                                                              \\ \hline
        BaseVISor                              & maintained               & usable                                                              \\ \hline
        BORN                                   & maintained               & usable                                                              \\ \hline
        Bossam                                 & abandoned                & no files to try available                                           \\ \hline
        CB (Consequence-based reasoner)        & abandoned                & no, compilation failed                                              \\ \hline
        CEL                                    & maintained               & usable                                                              \\ \hline
        Cerebra Engine                         & abandoned                & no files to try available                                           \\ \hline
        CICLOP                                 & abandoned                & no files to try available                                           \\ \hline
        Clipper                                & abandoned                & usable                                                              \\ \hline
        DBOWL                                  & abandoned                & no files to try available                                           \\ \hline
        Deslog                                 & abandoned                & no files to try available                                           \\ \hline
        DistEL                                 & abandoned                & no, missing info about how to compile it                            \\ \hline
        DLP                                    & abandoned                & no files to try available                                           \\ \hline
        DReW                                   & abandoned                & no, DLV required, but not available                                 \\ \hline
        ElepHant                               & maintained               & usable                                                              \\ \hline
        ELK                                    & maintained               & usable                                                              \\ \hline
        ELLY                                   & abandoned                & no, jar files seem broken                                           \\ \hline
        Expressive Reasoning Graph Store (IBM) & maintained               & usable                                                              \\ \hline
        EYE                                    & maintained               & usable                                                              \\ \hline
        EYE.js                                 & maintained               & usable                                                              \\ \hline
        F-OWL                                  & abandoned                & no files to try available                                           \\ \hline
        FaCT++                                 & abandoned                & usable                                                              \\ \hline
        Flower                                 & abandoned                & no, Maven build failed                                              \\ \hline
        fuzzyDL                                & abandoned                & not tried, because Protégé 4.3 didn‘t run                           \\ \hline
        HAM-ALC                                & abandoned                & no files to try available                                           \\ \hline
        HermiT                                 & abandoned                & usable                                                              \\ \hline
        HS-Reasoner                            & abandoned                & not tried, no knowledge about Haskell                               \\ \hline
        jcel                                   & maintained               & usable                                                              \\ \hline
        JFact                                  & maintained               & usable                                                              \\ \hline
        KAON2                                  & abandoned                & usable                                                              \\ \hline
        Konclude                               & maintained               & usable                                                              \\ \hline
        leanCoR                                & abandoned                & no, SWI setup in Ubuntu 20.04. failed                               \\ \hline
        LiFR                                   & maintained               & usable                                                              \\ \hline
        LillyTab                               & abandoned                & no, jar files seem broken                                           \\ \hline
        LiRoT                                  & maintained               & usable                                                              \\ \hline
        Mini-ME & maintained & no, download links for plugin broken [... ] \\ \hline
        Mini-ME Swift                          & no information available & usable                                                              \\ \hline
        NORA    & maintained & not tried, because it requires additional software [...]\\ \hline
        ODevice                                & abandoned                & not tried, because files are outdated \\ \hline
        Ontop (with Query reasoner)            & maintained               & usable                                                              \\ \hline
        Openllet                               & maintained               & usable                                                              \\ \hline
        Owlgres                                & abandoned                & no files to try available                                           \\ \hline
        OWLIM                                  & abandoned                & no files to try available                                           \\ \hline
        OWLRL                                  & maintained               & usable                                                              \\ \hline
        Pellet                                 & abandoned                & usable                                                              \\ \hline
        Pocket KRHyper                         & abandoned                & no files to try available                                           \\ \hline
        Pronto                                 & abandoned                & no, start script seems faulty                                       \\ \hline
        QueryPIE                               & abandoned                & not tried, because it requires additional software   \\ \hline
        Quill (part of TrOWL)                  & abandoned                & no files to try available                                           \\ \hline
        QuOnto                                 & abandoned                & no files to try available                                           \\ \hline
        RACER / RacerPro                       & abandoned                & no files to try available                                           \\ \hline
        RAT-OWL                                & abandoned                & no files to try available                                           \\ \hline
        RDFox                                  & maintained               & usable                                                              \\ \hline
        RDFSharp.Semantics                     & maintained               & usable                                                              \\ \hline
        reasonable                             & maintained               & usable                                                              \\ \hline
        REL (part of TrOWL)                    & abandoned                & no files to try available                                           \\ \hline
        Sequoia                                & maintained               & not tried, because it requires a Scala environment                  \\ \hline
        SHER                                   & abandoned                & no files to try available                                           \\ \hline
        Snorocket                              & abandoned                & no files to try available                                           \\ \hline
        SoftFacts                              & abandoned                & no files to try available                                           \\ \hline
        SparkEL                                & abandoned                & no, setup script seems faulty                                       \\ \hline
        SPOR                                   & abandoned                & not tried, no knowledge in compiling C++ programs                   \\ \hline
        SPOWL                                  & abandoned                & no files to try available                                           \\ \hline
        SWRL-IQ                                & abandoned                & no, outdated Protégé required                                       \\ \hline
        Tiny-ME                                & no information available & usable                                                              \\ \hline
        TReasoner                              & abandoned                & not tried, it requires a custom Java project   \\ \hline
        TRILL (and TRILLP and TORNADO)         & maintained               & usable                                                              \\ \hline
        Vampire                                & maintained               & usable                                                              \\ \hline
        VLog                                   & maintained               & usable                                                              \\ \hline
        WebPIE                                 & abandoned                & not tried, because it requires additional software      \\ \hline
        Whelk                                  & maintained               & usable                                                              \\ \hline
        Wolpertinger                           & abandoned                & no, Maven build failed                                              \\ \hline
    \end{tabular}
    \caption{List of all analyzed OWL reasoners \label{table-all-owl-reasoners}}
\end{table}

\begin{table}[]
    \begin{tabular}{|l|l|l|}
        \hline
        \textbf{Systems using foreign reasoners} & \textbf{Maintenance} & \textbf{Is usable}                                         \\ \hline
        AberOWL  & maintained & usable                                              \\ \hline
        BUNDLE   & maintained & usable                                              \\ \hline
        Chainsaw & maintained & not tried, because custom Java project required     \\ \hline
        ComR     & abandoned  & no files to try available                           \\ \hline
        COROR    & abandoned  & no files to try available                           \\ \hline
        DLEJena  & abandoned  & no files to try available                           \\ \hline
        DRAOn    & abandoned  & no files to try available                           \\ \hline
        ELOG     & abandoned  & usable                                              \\ \hline
        Hoolet   & abandoned  & usable                                              \\ \hline
        HydrOWL  & abandoned  & no files to try available                           \\ \hline
        HyLAR    & maintained & useable                                             \\ \hline
        Minerva  & abandoned  & no files to try available                           \\ \hline
        MORe     & abandoned  & not tried, because custom Java project required     \\ \hline
        NoHR     & abandoned  & useable                                             \\ \hline
        PAGOdA   & maintained & useable                                             \\ \hline
        PROSE    & abandoned  & no files to try available                           \\ \hline
        R2O2*    & abandoned  & no, compilation failed                              \\ \hline
        REQUIEM  & abandoned  & yes, Docker container was tried                     \\ \hline
        RuQAR    & abandoned  & no files to try available                           \\ \hline
        Screech  & abandoned  & no files to try available                           \\ \hline
        TrOWL    & abandoned  & not tried, I assume custom Java project is required \\ \hline
        WSClassifier (and WSReasoner)            & abandoned            & not tried, because files in various Google Drive folders \\ \hline
    \end{tabular}
    \caption{List of all analyzed systems using a third-party reasoner \label{table-all-systems}}
\end{table}

\medskip

\printbibliography

\end{document}